\documentclass{article}

\usepackage{PRIMEarxiv}

\usepackage[utf8]{inputenc} 
\usepackage[T1]{fontenc}    
\usepackage{hyperref}       
\usepackage{url}            
\usepackage{booktabs}       
\usepackage{amsfonts}       
\usepackage{nicefrac}       
\usepackage{microtype}      
\usepackage{lipsum}
\usepackage{fancyhdr}       
\usepackage{graphicx}       
\usepackage{natbib}
\usepackage{amsmath}
\usepackage{booktabs}
\usepackage{algorithm}
\usepackage{algpseudocode}
\usepackage{tikz}
\usetikzlibrary{arrows.meta,positioning,fit,backgrounds,shapes.geometric}
\usepackage{booktabs}
\usetikzlibrary{positioning,arrows.meta}
\graphicspath{{media/}}     

\title{Agentao: A Policy-Governed Runtime Harness for Embeddable Tool-Using LLM Agents
}

\author{
  Bo Jin\thanks{Corresponding author: jinbo@gass.cn} \\
  The Third Research Institute of the Ministry of Public Security‌  \\
  Shanghai\\
  \texttt{jinbo@gass.cn} \\
    \And
  Qiang Jiao \\
  Bureau of Science and Technology Information \\
  Ministry of Public Security of the People's Republic of China \\
  Beijing\\
  \texttt{jiaoqiangbj@126.com } 
  \AND
  Xin Tong \\
  School of Information and Cyber Security \\
  People’s Public Security University of China\\
  Beijing\\
  \texttt{tongxindotnet@outlook.com} \\
}

\begin{document}
\maketitle

\begin{abstract}
LLM agents increasingly operate as execution systems that invoke tools, modify local state, use persistent memory, and interact with external protocols. These capabilities make agents useful, but they also introduce risks related to over-privileged actions, weak auditability, prompt injection, tool poisoning, and uncontrolled side effects. This paper presents Agentao, a governed local-first runtime for tool-using LLM agents. Agentao separates model-generated action proposals from host-authorized execution through a layered architecture consisting of host-facing surfaces, a host contract, a runtime core, a permission-mediated tool system, and supporting subsystems for memory, replay, plugins, skills, sub-agents, and protocol integration. We describe the motivation, threat model, design goals, governance model, execution pipeline, and structured event interface of the system. Agentao does not provide formal safety guarantees; rather, it demonstrates how permissions, state, protocol boundaries, and execution traces can be made explicit runtime abstractions for building agents that are more governable, inspectable, and suitable for host-controlled local environments. The code is publicly available at \url{https://github.com/jin-bo/agentao}.
\end{abstract}


\section{Introduction}

Large language model (LLM) agents are increasingly moving beyond single-turn interaction and static text generation toward systems that plan, invoke tools, coordinate with external services, maintain state, and operate across multiple steps. This transition changes the central research problem. Once an LLM system is allowed to inspect files, call APIs, execute commands, or interact with other agents, the key challenge is no longer merely whether the model can choose useful actions, but whether the surrounding runtime can make such actions governable, observable, recoverable, and interoperable. In this sense, agent systems are not only model-centered applications; they are socio-technical execution environments in which autonomy, authority, state, and accountability must be jointly engineered.

Existing agent frameworks and protocols address different parts of this emerging stack. Orchestration frameworks emphasize durable execution, stateful workflows, streaming, human-in-the-loop intervention, and long-running agent control. Agent SDKs expose code-first abstractions for applications that own tool execution, orchestration, approvals, and state. Protocols such as the Model Context Protocol (MCP) standardize how LLM applications connect to external data sources and tools, while the Agent Client Protocol (ACP) standardizes communication between code editors or IDEs and coding agents. These developments indicate a broader shift from isolated tool-calling loops toward open, composable agent ecosystems. However, protocol-level connectivity and framework-level orchestration do not by themselves provide a complete answer to runtime governance: who authorizes an action, what boundary constrains it, how the decision is recorded, and how the resulting execution can be inspected or replayed remain system-level questions.

Agentao is designed within this design space. Rather than treating an agent as a monolithic assistant or a collection of ad hoc tool calls, Agentao presents a local-first, private-first, embeddable runtime in which permissions, protocols, memory, plugins, and multi-session control are treated as first-class concerns. The system is therefore best understood not as yet another agent interface, but as a governed runtime boundary between a host application and the capabilities exposed to an LLM-driven agent. This boundary is important because the same agent may interact with local files, external tools, persistent memory, remote protocol endpoints, and auxiliary agents. Without a coherent runtime contract, such capabilities risk becoming fragmented across separate libraries, user interfaces, configuration files, and implicit trust assumptions.

This paper studies Agentao as an instance of a governed agent runtime. We use the term \emph{governed runtime} to refer to a runtime environment that mediates agent actions through explicit policy, exposes execution state to the host, records relevant decisions and outcomes, and composes external capabilities without surrendering host-side control. This framing deliberately avoids claiming that governance can be solved solely by prompting or by model alignment. Instead, it treats governance as a runtime property: an agent action should be evaluated in relation to permission modes, filesystem or network boundaries, tool provenance, user consent, session state, and auditability. The resulting research question is:

\begin{quote}
How can a local-first agent runtime provide useful autonomy while preserving host control, permission-mediated execution, protocol interoperability, and auditable system behavior?
\end{quote}

This question is motivated by the need to bound agent autonomy, compose heterogeneous tool protocols under a common authority model, and preserve structured evidence of runtime behavior.

Agentao addresses these concerns through a runtime architecture organized around governance, connectivity, and observability. Governance concerns the enforcement of permission modes and consent boundaries before actions are executed. Connectivity concerns the controlled composition of local tools, protocol-mediated tools, plugins, memory, and auxiliary agents. Observability concerns the production of structured execution traces that allow host applications and users to inspect, debug, and validate agent behavior. These three concerns are mutually reinforcing: permission decisions are more useful when they are observable; protocol integration is safer when routed through a common governance layer; and memory or sub-agent execution is more manageable when captured by a unified runtime contract.

The contributions of this paper are threefold. 

First, we formulate governed local-first runtime design as a systems problem distinct from prompting, orchestration, and protocol connectivity alone. 

Second, we present Agentao as a layered runtime architecture that separates model-generated action proposals from host-authorized execution through a host contract, runtime core, permission-mediated tool pipeline, and supporting subsystems. 

Third, we describe the mechanisms that make agent execution more governable and inspectable, including permission modes, confirmation flows, sandbox-aware execution, scoped memory, protocol-mediated capability registration, structured host events, session persistence, and replayable traces.

The purpose of this paper is not to argue that Agentao eliminates all risks associated with tool-using LLM agents. It does not provide a formal proof of safety, nor does it remove the need for secure tool implementations, careful deployment policies, or robust model-level defenses. Instead, the paper advances a more modest systems claim: a local-first agent runtime can make agent behavior more governable and inspectable by making permissions, protocol boundaries, execution state, and audit traces explicit parts of the runtime architecture.

\section{Related Work}
\label{sec:related-work}

\subsection{Tool-Augmented Language Models and Acting Agents}

Early work on tool-augmented language models established that the capabilities of large language models can be extended by coupling generation with external computation, retrieval, or environment interaction. MRKL systems proposed a modular neuro-symbolic architecture in which a language model routes subproblems to specialized symbolic or neural modules, thereby framing tool use as a systems problem rather than merely a prompting strategy \citep{karpas2022mrkl}. Toolformer showed that language models can be trained to decide when and how to call external APIs using self-supervised signals, including tools such as calculators, search engines, translation systems, and calendars \citep{schick2023toolformer}. ReAct further connected reasoning and acting by interleaving natural-language reasoning traces with environment actions, demonstrating improved interpretability and task performance in knowledge-intensive and interactive decision-making settings \citep{yao2022react}.

Subsequent work broadened tool use from a small number of APIs to large-scale tool ecosystems. API-Bank introduced a benchmark and training corpus for tool-augmented dialogue, emphasizing planning, tool retrieval, and API calling as separable evaluation dimensions \citep{li2023apibank}. Gorilla studied API invocation at scale and showed that retrieval over documentation can improve model robustness to changing API specifications \citep{patil2023gorilla}. HuggingGPT and TaskMatrix.AI framed foundation models as controllers that decompose user requests and dispatch subtasks to external models or APIs \citep{shen2023hugginggpt,liang2023taskmatrix}. Together, these works motivate the view that modern LLM applications are increasingly mediated by external capabilities. However, they primarily focus on action selection, tool correctness, and task completion; comparatively less attention is given to the runtime boundary that authorizes, constrains, records, and audits tool invocations.

\subsection{Language-Agent Architectures, Memory, and Long-Horizon Autonomy}

A second line of work studies language agents as cognitive or computational architectures with explicit modules for memory, planning, reflection, and action. CoALA proposes a cognitive architecture for language agents, organizing existing systems around modular memory components, structured action spaces, and generalized decision-making procedures \citep{sumers2023coala}. This perspective is useful because it treats an agent not as a single model call but as an arrangement of control loops, state representations, and environment interfaces. Surveys of LLM-based autonomous agents similarly identify perception, memory, planning, and action as recurring architectural components across agent systems \citep{wang2023surveyagents,xi2023riseagents}.

Long-horizon autonomy has also been studied through memory and self-improvement mechanisms. Generative Agents introduced an architecture in which agents store observations, retrieve relevant memories, and synthesize reflections to support believable social behavior in an interactive simulation \citep{park2023generativeagents}. Reflexion uses verbal feedback stored in episodic memory to improve future decision-making without updating model weights \citep{shinn2023reflexion}. Voyager studies open-ended embodied learning in Minecraft, using an automatic curriculum, iterative prompting, and a skill library of executable code to accumulate reusable behaviors over time \citep{wang2023voyager}. Language Agent Tree Search integrates planning, acting, and reflection through a tree-search procedure over possible agent trajectories \citep{zhou2023lats}. These systems show that memory and reusable skills can substantially extend agent capability. The present work is complementary: rather than optimizing memory or planning alone, it emphasizes the runtime conditions under which persistent state, reusable skills, and external actions can be made governable and inspectable.

\subsection{Agent Frameworks, Multi-Agent Systems, and Runtime Infrastructure}

Recent research has also investigated software infrastructures for composing, coordinating, and executing LLM agents. AutoGen formulates multi-agent applications as conversations among customizable agents that may combine LLM reasoning, human input, and tool execution \citep{wu2023autogen}. Broader surveys of LLM-based multi-agent systems highlight communication, role assignment, collaboration protocols, and emergent coordination as central design questions \citep{guo2024multiagents}. These works demonstrate the importance of agent composition, but their primary abstraction is often conversational coordination rather than host-governed execution.

Other work approaches agents from the perspective of computing environments and operating-system-like services. OS-Copilot studies generalist computer agents that interact with web browsers, terminals, files, and applications, emphasizing self-improvement across heterogeneous operating-system tasks \citep{wu2024oscopilot}. SWE-agent introduces the notion of an agent-computer interface for software engineering agents and shows that interface design can materially affect an agent's ability to navigate repositories, edit files, and run tests \citep{yang2024sweagent}. AIOS proposes an operating-system architecture for LLM agents, separating agent applications from kernel-level services such as scheduling, context management, memory management, storage management, and access control \citep{mei2024aios}. These systems are closely related in their recognition that LLM agents require explicit runtime support. The distinction made here is to treat governance, auditability, and permission-mediated action as first-class runtime properties rather than as secondary implementation concerns.

\subsection{Benchmarks and Evaluation of Digital Agents}

The evaluation literature provides evidence that tool-using agents remain brittle in realistic digital environments. WebArena constructs a reproducible web environment with functional websites and long-horizon tasks, showing a substantial gap between human performance and GPT-4-based agents on realistic web tasks \citep{zhou2023webarena}. GAIA evaluates general AI assistants on questions requiring reasoning, multimodality, browsing, and tool-use proficiency, again exposing a large gap between human robustness and model performance \citep{mialon2023gaia}. SWE-bench evaluates whether language models can resolve real GitHub issues by editing codebases, revealing that real-world software engineering tasks require long-context understanding, execution feedback, and coordinated edits across files \citep{jimenez2023swebench}. These benchmarks are important because they shift evaluation away from isolated text generation toward situated execution.

However, standard agent benchmarks typically measure task success, final-answer correctness, or issue-resolution rates. They do not fully capture whether an agent remained within an intended authority boundary, whether risky actions were surfaced for confirmation, whether a host can reconstruct the execution trace, or whether tool-mediated state changes are auditable after the fact. Safety-oriented benchmarks begin to address these concerns. Agent-SafetyBench evaluates LLM agents across multiple safety-risk categories and reports that existing agents remain vulnerable to unsafe interaction patterns \citep{zhang2024agentsafetybench}. AgentHarm focuses on harmful multi-step agent tasks and shows that agentic misuse can remain coherent even under jailbreak conditions \citep{andriushchenko2024agentharm}. Agent Security Bench formalizes attacks and defenses against LLM-based agents across scenarios involving tools, memory, and prompt handling \citep{zhang2024asb}. These results motivate runtime designs that expose authority boundaries and execution traces, even when the primary contribution is architectural rather than benchmark-driven.

\subsection{Prompt Injection, Tool Poisoning, and Agent Security}

Security research has shown that LLM-integrated applications blur the boundary between data and instructions. Indirect prompt injection attacks demonstrate that malicious instructions embedded in retrieved or external content can manipulate LLM-integrated applications, including systems that call APIs or process untrusted web content \citep{greshake2023notwhat}. Other work studies prompt injection against real-world LLM-integrated applications and formalizes attack and defense settings for systematic evaluation \citep{liu2023promptinjection,liu2023formalizing}. These findings are especially relevant for agents because tool use increases the consequences of model confusion: a compromised instruction may no longer merely alter text output, but may trigger file operations, API calls, database updates, or cross-system data transfer.

Recent work extends these concerns to agent-specific and protocol-mediated tool ecosystems. Research on MCP-integrated systems identifies tool poisoning, shadowing, and descriptor manipulation as attacks in which malicious instructions or semantic cues are embedded in tool metadata rather than ordinary user prompts \citep{hou2025model}. Threat-modeling work on MCP highlights client-side vulnerabilities, including prompt injection through tool metadata and insufficient transparency around tool parameters and model decision paths \citep{huang2026mcptm}. MCP-SafetyBench evaluates LLMs interacting with real-world MCP servers and reports that multi-server and multi-turn workflows introduce safety risks not captured by isolated prompt-injection tests \citep{zong2025mcpsafetybench}. More generally, work on verifiably safe tool use argues that model-based safeguards alone are insufficient for high-stakes tool-using agents and proposes deriving enforceable specifications over data flows and tool sequences \citep{doshi2026safetooluse}. This literature motivates runtime-level mechanisms for least privilege, confirmation, provenance, isolation, and auditability.

\subsection{Positioning}

The above literature establishes the technical foundations for tool-using and long-horizon LLM agents, but it also reveals a gap. Tool-use research improves the ability of models to select and invoke capabilities; agent-architecture research improves planning, memory, and skill reuse; multi-agent frameworks improve coordination; and safety benchmarks identify vulnerabilities in prompt handling, memory, and tool execution. Yet a practical agent system also requires a runtime boundary that mediates authority between the host, the user, the model, and external capabilities. The present work is positioned at this boundary. Its focus is not to propose a new prompting method, a new model, or a new benchmark, but to articulate and instantiate a governed runtime design in which permissions, protocol composition, state, and observability are explicit system abstractions.

Table~\ref{tab:positioning} positions Agentao relative to representative
agent frameworks, runtimes, coding agents, and interoperability
protocols. The comparison is qualitative and is not intended to rank
systems by task performance. Instead, it clarifies that Agentao targets
the runtime harness layer: the boundary at which a host application
authorizes, executes, observes, and replays model-proposed actions.

\begin{table}[t]
\centering
\small
\caption{Qualitative positioning of Agentao relative to representative
agent frameworks, runtimes, coding agents, and interoperability protocols.
The comparison focuses on system abstractions rather than benchmark
performance.}
\begin{tabular}{p{0.18\linewidth} p{0.31\linewidth} p{0.41\linewidth}}
\hline
\textbf{System} & \textbf{Primary abstraction} & \textbf{Relation to Agentao} \\
\hline

LangChain / LangGraph &
Agent framework and orchestration runtime. LangChain provides
abstractions for models, tools, agent loops, and middleware, while
LangGraph provides low-level orchestration for long-running, stateful
agents with durable execution, streaming, human-in-the-loop control, and
persistence. &
Agentao is complementary to graph-based orchestration. Its focus is the
host-controlled runtime harness: permission-mediated tool execution,
confirmation, scoped local state, and host-facing audit events before
model-proposed actions affect local or external resources. \\

LlamaIndex &
Data-centric framework for context-augmented LLM applications, including
data connectors, indexes, query engines, chat engines, agents, workflows,
and observability/evaluation integrations. &
Agentao can use retrieval, database, or document-processing tools as
capabilities, but its primary abstraction is not an index or RAG
pipeline. It focuses on governing executable actions over local files,
shell commands, memory, external tools, and protocol-mediated
capabilities. \\

AutoGen &
Framework for building single-agent and multi-agent applications,
including conversational agent patterns, event-driven multi-agent
systems, extensions, MCP workbenches, and code-execution components. &
Agentao supports sub-agents and tool execution, but treats delegation and
tool use as host-visible runtime events subject to permission policy,
confirmation, and replay rather than primarily as conversational
coordination among agents. \\

CrewAI &
Multi-agent automation framework centered on agents, crews, flows,
tasks, processes, guardrails, memory, knowledge, callbacks, and
observability for collaborative agent workflows. &
Agentao overlaps with CrewAI in supporting tools, memory, and
application workflows, but emphasizes a lower-level governed execution
boundary: active permission modes, policy-mediated tool calls,
sandbox-aware execution, and structured host contracts for embedding. \\

OpenAI Agents SDK / Google ADK &
Code-first agent SDKs for building production agent applications with
tools, handoffs or multi-agent workflows, sessions, tracing,
guardrails, sandboxed or managed execution, and deployment-oriented
runtime support. &
Agentao is similar in treating agents as managed runtimes, but is
designed as a local-first embeddable harness with explicit host
surfaces, replaceable local capabilities, permission snapshots,
policy-gated side effects, and replayable host-facing traces. \\

Coding-agent systems
(e.g., SWE-agent, OpenHands) &
Task-specialized agents for software engineering workflows such as
repository editing, issue fixing, terminal use, code review, and
developer automation. &
Agentao shares the concern for local files and shell-mediated actions,
but generalizes beyond coding tasks. It aims to provide a reusable
runtime harness that can be embedded in SaaS assistants, data workbenches,
batch jobs, audit pipelines, ACP servers, and other host applications. \\

Agent protocols
(MCP / ACP) &
Interoperability protocols. MCP standardizes how AI applications connect
to external data sources, tools, and workflows; ACP standardizes
communication between code editors or IDEs and coding agents. &
Agentao uses protocols as connectivity mechanisms, but does not treat
protocol connectivity as sufficient governance. Protocol-derived tools
are normalized into the same tool system and routed through the same
permission, confirmation, and event-trace mechanisms as local tools. \\

Agentao &
Policy-governed runtime harness for embeddable tool-using LLM agents. &
Agentao's distinguishing focus is host-controlled execution: permission
modes, policy-mediated tool invocation, confirmation flows, hardline
safety checks, sandbox-aware execution, scoped memory, structured host
events, session persistence, and replayable traces. \\
\hline
\end{tabular}

\label{tab:positioning}
\end{table}

\section{Motivation, Threat Model, and Design Goals}
\label{sec:design-goals}

This section presents the motivation, threat model, and design goals that guide the design of a governed local-first agent runtime. The central premise is that LLM agents should not be treated merely as text-generation systems with tool access, but as execution systems whose actions may affect local state, external services, persistent memory, and user-controlled resources. Consequently, agent autonomy must be considered together with authority, accountability, and recoverability.

\subsection{Motivation}
\label{subsec:motivation}

LLM agents differ from conventional chatbots because they can produce side effects: they may inspect or modify files, invoke APIs, run commands, access memory, and coordinate with auxiliary agents. In a local-first setting, these actions may affect sensitive resources such as source repositories, credentials, configuration files, personal documents, and project-specific memory. The runtime must therefore determine not only whether an action is useful, but whether it is authorized, bounded, and inspectable.

Protocol composability further complicates this problem. Tools may originate from local implementations, plugins, protocol servers, skills, or sub-agents, each with different provenance and side-effect semantics. A governed runtime is motivated by the need to mediate these heterogeneous capabilities through a common authority boundary and to record enough structured evidence for debugging, review, and accountability.

\subsection{Threat Model}
\label{subsec:threat-model}

We consider an agent runtime in which an LLM-driven agent can interact with local resources, external tools, persistent memory, plugins, protocol-mediated capabilities, and auxiliary agents. The runtime is embedded in a host application or exposed through an interactive interface. The user may delegate tasks to the agent, but does not necessarily intend to grant unrestricted authority over all available resources.

\paragraph{Assets.}
The primary assets are: (i) local files and directories, including source code, documents, configuration files, and generated artifacts; (ii) credentials, tokens, environment variables, and other secrets; (iii) persistent memory and session state; (iv) external services reachable through tools or protocols; (v) the integrity of tool outputs and execution results; and (vi) audit logs or traces used for accountability. The runtime should preserve the confidentiality, integrity, and availability of these assets to the extent possible under the host's policy.

\paragraph{Adversaries and failure sources.}
We distinguish between malicious actors and unreliable components. Adversarial inputs may originate from a user prompt, an external document, a webpage, a tool response, a plugin, a protocol server, a repository, or another agent. A tool provider or protocol endpoint may be benign, compromised, misconfigured, or malicious. The LLM itself is not modeled as a malicious principal, but as an unreliable decision-making component that may be manipulated by adversarial instructions, may misunderstand policy, may overgeneralize from context, or may select unsafe actions.

\paragraph{Attack surfaces.}
Table~\ref{tab:threat-model} summarizes representative threats. The list is not intended to be exhaustive; rather, it identifies the classes of risk most relevant to a local-first, tool-using agent runtime.

\begin{table}[t]
\centering
\caption{Representative threats for a governed local-first agent runtime.}
\small
\begin{tabular}{p{0.22\linewidth}p{0.36\linewidth}p{0.32\linewidth}}
\toprule
\textbf{Threat} & \textbf{Example} & \textbf{Runtime concern} \\
\midrule
Over-privileged action & The agent modifies files outside the user's intended task scope. & Permission boundaries and least privilege. \\
Prompt injection & An external document instructs the model to ignore prior instructions or exfiltrate data. & Separation of data from authority and policy-mediated action. \\
Tool poisoning & A tool description or metadata field contains hidden instructions that alter model behavior. & Tool provenance, metadata validation, and user-visible capability review. \\
Confused deputy & The model uses a legitimate capability to perform an action for an unintended principal or purpose. & Context-sensitive authorization and consent. \\
State leakage & Persistent memory or session context exposes sensitive information in later tasks. & Memory scoping, redaction, and state lifecycle management. \\
Irreversible side effect & A command deletes, overwrites, sends, publishes, or otherwise commits state. & Confirmation, cancellation where supported, replay-based diagnosis, and fail-closed execution. \\
Repudiation & After a failure, the host cannot determine which tool call or decision caused the issue. & Structured audit events and replayable traces. \\
Extension supply-chain risk & A plugin, skill, protocol server, or sub-agent expands the action space beyond what the user expected. & Capability registration, isolation, and policy-aware composition. \\
\bottomrule
\end{tabular}
\label{tab:threat-model}
\end{table}

\paragraph{Trust assumptions.}
We assume that the host application and the runtime enforcement mechanism are part of the trusted computing base. The operating system, filesystem permissions, and process isolation mechanisms are assumed to behave according to their documented semantics. We do not assume that tool outputs, external content, plugin metadata, protocol server descriptions, or model-generated plans are trustworthy. We also do not assume that natural-language instructions are a reliable security boundary.

\paragraph{Non-goals.}
The runtime does not attempt to solve all LLM safety problems. It does not provide a formal proof that no unsafe behavior can occur. It does not defend against a fully compromised host system, kernel, or operating-system account. It does not address model training-time attacks, model extraction, or general jailbreak resistance as standalone problems. It also cannot guarantee that an external service invoked by a tool will behave correctly once a request has been authorized. The goal is narrower: to reduce the authority of the model-mediated execution path, make risky actions explicit, and provide structured evidence about what occurred.

\subsection{Design Goals}
\label{subsec:design-goals}

The threat model leads to five design goals. These goals are intended to guide governed agent runtimes that operate over local resources, external tools, persistent state, and protocol-mediated capabilities.

\paragraph{G1: Bounded and consent-mediated autonomy.}
The runtime should support useful agent autonomy without equating autonomy with unrestricted execution. Different tasks require different levels of authority: some can be completed through read-only inspection, others require project-scoped modification, and high-impact operations may require explicit user or host confirmation. Agent autonomy should therefore be expressed through permission modes, execution boundaries, and consent flows rather than through implicit trust in model-generated actions.

\paragraph{G2: Least-privilege capability exposure.}
The runtime should expose and authorize capabilities according to task scope, operation type, and possible side effects. File access, shell execution, network operations, memory updates, protocol-mediated tools, and sub-agent delegation should not be treated as equivalent tool calls. Each capability should be constrained according to the sensitivity of the resources it can access and the consequences it can produce.

\paragraph{G3: Protocol-composable governance.}
External protocols, plugins, skills, and auxiliary agents should be integrated through a common governance boundary. A capability should not bypass permission checks simply because it originates from an external server, plugin, or sub-agent. Conversely, the runtime should not require every integration to be rewritten as a privileged internal tool. The design should allow heterogeneous capabilities to be registered, described, invoked, constrained, and audited under a unified policy model.

\paragraph{G4: State and memory discipline.}
Persistent state can improve long-horizon assistance, but it also creates risks of leakage, stale context, and unintended cross-session influence. The runtime should distinguish transient context, session state, persistent memory, and generated artifacts. Memory access should be scoped and observable, and long-term state should support continuity without silently expanding the model's authority.

\paragraph{G5: Structured auditability and recoverability.}
The runtime should produce structured records of important execution events, including proposed actions, permission decisions, tool invocations, errors, state changes, and sub-agent activity. These records should support debugging, replay, policy review, and post-hoc incident analysis. A local-first runtime should fail closed when authority is ambiguous and should preserve enough execution evidence for the host or user to reconstruct what happened.

Together, these goals define the runtime-level responsibilities required when LLM agents are allowed to act over local resources, external tools, persistent state, and protocol-mediated capabilities. They do not eliminate the need for secure tools, robust models, or careful deployment policies, but they make autonomy, authority, and accountability explicit parts of the system design.

\section{System Architecture}
\label{sec:system-architecture}

This section describes Agentao as a layered runtime architecture that separates model-generated action proposals from host-authorized execution. We focus on the host contract, runtime core, governed tool pipeline, protocol-composable tool system, supporting subsystems, and structured observability interface.

\subsection{Architectural Overview}
\label{subsec:architecture-overview}

\begin{figure*}[t]
    \centering
    \includegraphics[width=\textwidth]{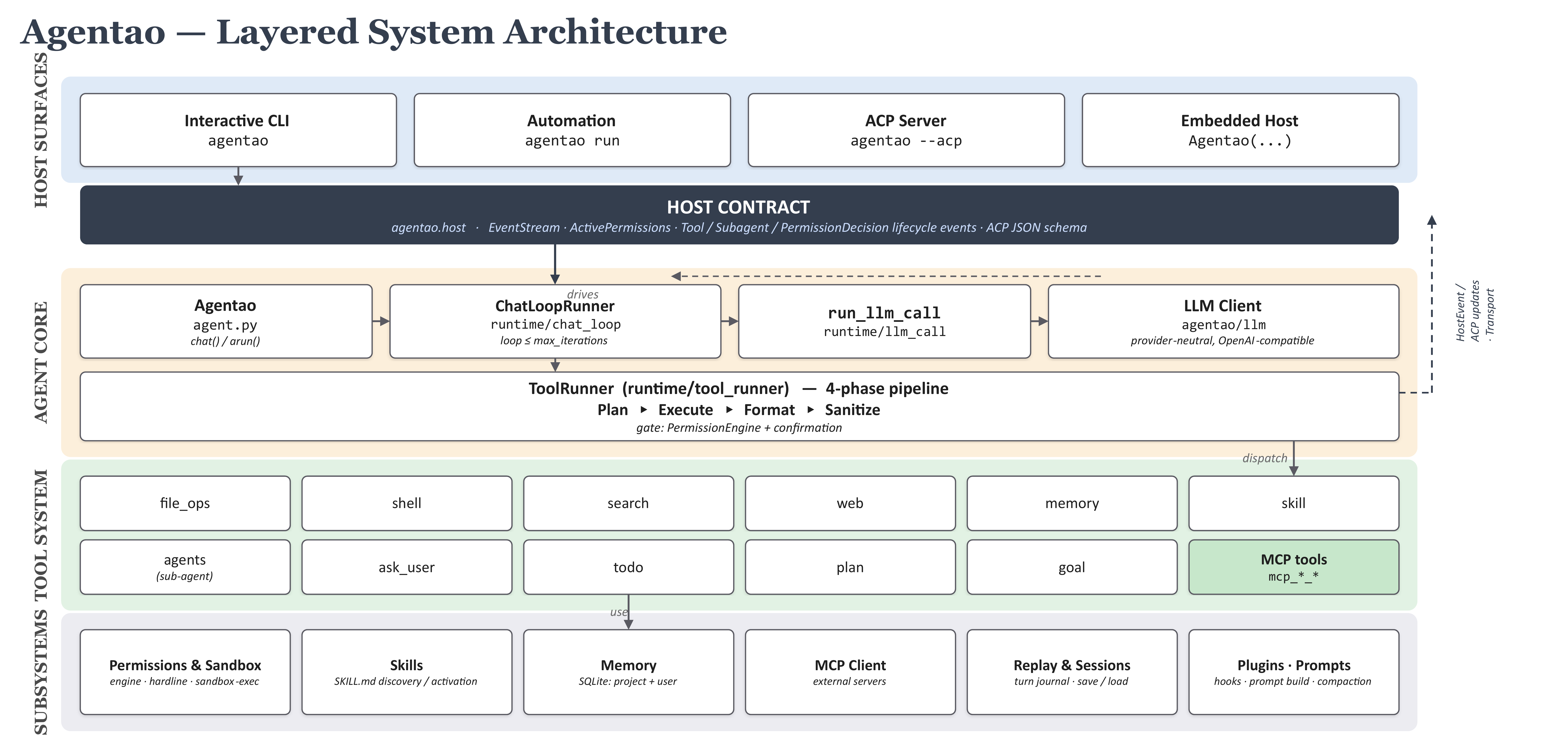}
    \caption{
    Layered system architecture of Agentao. Host-facing surfaces share a common host contract; the agent core coordinates model interaction and governed tool execution; the tool system exposes local and protocol-mediated capabilities; and supporting subsystems provide permissions, sandboxing, memory, replay, sessions, plugins, prompts, and protocol integration.
    }
    \label{fig:agentao-architecture}
\end{figure*}

Agentao is organized as a layered runtime architecture, as shown in Figure~\ref{fig:agentao-architecture}. The architecture separates host-facing interaction, runtime coordination, governed tool execution, and supporting subsystems. This separation is central to the system design: model-generated actions are not executed directly against local or external resources, but are routed through a host contract and a permission-mediated tool pipeline.

The top layer consists of \emph{host surfaces}. Agentao can be exposed through an interactive command-line interface, an automation-oriented execution mode, an ACP server mode, or an embedded host API. These surfaces differ in how users or applications initiate tasks, but they share the same underlying host contract. The host contract defines the runtime interface exposed to embedding applications, including event streams, active permissions, tool lifecycle events, sub-agent lifecycle events, permission-decision events, and protocol schemas.

The agent core implements the main execution loop. A user request enters the runtime through the Agentao interface and is processed by the chat-loop runner. The runner constructs model context, invokes the LLM client through a provider-neutral interface, receives assistant messages and tool-call proposals, and delegates executable actions to the tool runner. The tool runner implements a four-phase pipeline: \emph{plan}, \emph{execute}, \emph{format}, and \emph{sanitize}. The planning phase normalizes tool calls and prepares execution plans; the execution phase applies permission checks and dispatches approved calls; the formatting phase converts tool results into model-readable messages; and the sanitization phase redacts or normalizes outputs before they are exposed to the model or host.

The tool system contains the runtime-visible capabilities available to the agent. These include local file operations, shell execution, search, web access, memory operations, skills, sub-agents, user queries, task planning, goal management, and protocol-mediated MCP tools. Although these capabilities are presented to the model as callable tools, they are not granted uniform authority. Each invocation is mediated by the permission engine and may be allowed, denied, or routed to confirmation depending on the active mode, tool metadata, arguments, and host policy.

The bottom layer contains supporting subsystems. These include the permissions and sandbox subsystem, skill discovery and activation, scoped memory, the MCP client, replay and session persistence, and plugin or prompt hooks. These subsystems are not merely auxiliary implementation details. They provide the mechanisms through which Agentao realizes local-first governance, protocol composition, persistent state, extension, and auditability.

The architecture can be summarized as a runtime tuple
\begin{equation}
    \mathcal{R} =
    \langle H, C, G, T, S, E \rangle ,
\end{equation}
where $H$ denotes host surfaces, $C$ the host contract, $G$ the agent core, $T$ the tool system, $S$ the supporting subsystems, and $E$ the structured event stream. A model-generated action is coordinated by $G$ and mediated by the governed tool pipeline before reaching $T$, while $C$ exposes the corresponding permissions, decisions, and execution events to the host through $E$.

\subsection{Host Contract and Runtime Core}
\label{subsec:host-contract-runtime-core}

The host contract is the boundary between Agentao and the environment in which it is embedded. It defines what the host can observe, control, and configure. Rather than exposing only a final assistant message, the contract exposes active permissions, tool lifecycle events, sub-agent lifecycle events, permission-decision events, transport updates, and protocol schemas. This makes the runtime suitable not only for interactive use, but also for automation, editor integration, and embedding in larger applications.

Formally, a host interaction can be represented as:
\begin{equation}
    h_t = \langle u_t, \pi_t, \omega_t \rangle ,
\end{equation}
where $u_t$ is the user or host input at turn $t$, $\pi_t$ is the active permission context, and $\omega_t$ is the host-observable event stream prefix. The runtime transforms the interaction state according to:
\begin{equation}
    \Sigma_{t+1}, \tau_t, r_t =
    \mathrm{RunTurn}(\Sigma_t, h_t, M, T),
\end{equation}
where $\Sigma_t$ is runtime state, $M$ is the LLM client, $T$ is the tool system, $\tau_t$ is the trace generated during the turn, and $r_t$ is the final response or terminal status.

The runtime core is responsible for context construction, model invocation, tool-call normalization, iteration control, cancellation, and result integration. Each turn begins by building the model input from the user request, system instructions, conversation history, available tool schemas, relevant memory, plan state, and host-provided constraints. The LLM then returns either a final response or one or more tool-call proposals.

A normalized tool proposal has the form:
\begin{equation}
    p = \langle id, n, a, c, m \rangle ,
\end{equation}
where $id$ is a stable tool-call identifier, $n$ is the tool name, $a$ is the decoded argument object, $c \in T$ is the resolved capability, and $m$ is metadata used for permission checking and event correlation. Stable identifiers are important because they allow permission decisions, lifecycle events, tool outputs, and replay records to be correlated even when model providers differ in how they represent tool calls.

Algorithm~\ref{alg:governed-turn} summarizes the governed turn loop. The key property is that tool selection and tool execution are separated. The model may propose an action, but the runtime determines whether, how, and under what constraints the action is executed.

\begin{algorithm}[t]
\caption{Governed agent turn}
\label{alg:governed-turn}
\begin{algorithmic}[1]
\Require user input $u$, runtime state $\Sigma$, model $M$, tool system $T$, permission policy $\mathcal{P}$, host contract $C$
\Ensure updated state $\Sigma'$, final response $r$, trace $\tau$
\State $\tau \gets [\ ]$
\State $x \gets \Call{BuildContext}{u,\Sigma,T}$
\For{$k = 1$ to $K_{\max}$}
    \State $y \gets M(x)$
    \If{$y$ contains no tool calls}
        \State $r \gets \Call{ExtractFinalResponse}{y}$
        \State \Return $\Sigma, r, \tau$
    \EndIf

    \State $Q \gets \Call{NormalizeToolCalls}{y,T}$
    \State $A \gets [\ ]$

    \ForAll{$p \in Q$}
        \If{\Call{InvalidProposal}{$p$}}
            \State $x \gets x \oplus \Call{ToolErrorMessage}{p}$
            \State \textbf{continue}
        \EndIf

        \State $d \gets \Call{Decide}{p,\mathcal{P},\Sigma}$
        \State $\tau \gets \tau \oplus \Call{PermissionEvent}{p,d}$

        \If{$d=\textsc{deny}$}
            \State $x \gets x \oplus \Call{DeniedToolMessage}{p}$
        \ElsIf{$d=\textsc{prompt}$}
            \State $b \gets C.\Call{Confirm}{p}$
            \If{$b=\mathrm{true}$}
                \State $A \gets A \oplus \{p\}$
            \Else
                \State $x \gets x \oplus \Call{CancelledToolMessage}{p}$
            \EndIf
        \Else
            \State $A \gets A \oplus \{p\}$
        \EndIf
    \EndFor

    \State $R \gets \Call{ExecuteAllowed}{A,\Sigma}$
    \State $\tau \gets \tau \oplus \Call{LifecycleEvents}{A,R}$
    \State $x \gets x \oplus \Call{FormatAndSanitize}{R}$
    \State $\Sigma \gets \Call{UpdateState}{\Sigma,R}$
\EndFor
\State \Return $\Sigma, \Call{IterationLimitResponse}{}, \tau$
\end{algorithmic}
\end{algorithm}

\subsection{Governed Tool Execution}
\label{subsec:governed-tool-execution}

Governance in Agentao is implemented as a cross-cutting runtime function rather than as a standalone tool. It appears at three points in the layered architecture: the host contract exposes active permissions and permission-decision events; the ToolRunner pipeline gates tool execution through permission checking and confirmation; and the permissions and sandbox subsystem provides lower-level enforcement for sensitive operations such as shell execution and filesystem access.

The permission decision function maps a proposed tool invocation to one of three outcomes:
\begin{equation}
    D: T \times A \times \Sigma \times \mathcal{P}
    \rightarrow
    \{\textsc{allow}, \textsc{prompt}, \textsc{deny}\},
\end{equation}
where $T$ is the tool system, $A$ is the argument space, $\Sigma$ is runtime state, and $\mathcal{P}$ is the active permission policy.

The policy $\mathcal{P}$ is composed from multiple sources: hardline safety checks, per-run restrictions, user or host rules, permission-mode presets, and tool-level defaults. This layered structure is deliberate. Some operations should be denied before ordinary rule evaluation. Some restrictions are task-specific and should not be shadowed by broader allow rules. Other decisions depend on the active mode selected by the user or host.

Table~\ref{tab:permission-modes} summarizes the coarse permission modes.

\begin{table}[t]
\centering
\small
\begin{tabular}{p{0.22\linewidth}p{0.32\linewidth}p{0.34\linewidth}}
\toprule
\textbf{Mode} & \textbf{Intended authority} & \textbf{Typical behavior} \\
\midrule
\texttt{read-only} & Inspection without mutation. & Deny non-read-only capabilities; allow safe reading and search. \\
\texttt{workspace-write} & Project-scoped modification. & Allow controlled file edits; prompt for broad shell or network actions; deny destructive commands. \\
\texttt{full-access} & User-delegated broad operation. & Allow ordinary capabilities while preserving fail-closed checks for high-risk actions. \\
\texttt{plan} & Deliberation without side effects. & Allow planning and safe inspection; deny file writes, memory writes, and task mutation. \\
\bottomrule
\end{tabular}
\caption{Permission modes as coarse autonomy levels. Rule evaluation and capability metadata refine these presets into concrete decisions for individual tool calls.}
\label{tab:permission-modes}
\end{table}

A policy rule has the abstract form:
\begin{equation}
    r = \langle q_n, q_a, q_s, o \rangle ,
\end{equation}
where $q_n$ is a predicate over tool names, $q_a$ is a predicate over arguments, $q_s$ is a predicate over runtime state, and $o \in \{\textsc{allow}, \textsc{prompt}, \textsc{deny}\}$ is the rule outcome. A rule matches a proposed tool call $p$ when:
\begin{equation}
    \mathrm{match}(r,p,\Sigma)
    =
    q_n(p.n) \wedge q_a(p.a) \wedge q_s(\Sigma).
\end{equation}

Algorithm~\ref{alg:permission-decision} gives the permission decision procedure. The procedure is conservative: malformed calls, unknown tools, hardline-unsafe actions, and ambiguous authority boundaries fail closed rather than being silently executed.

\begin{algorithm}[t]
\caption{Permission decision}
\label{alg:permission-decision}
\begin{algorithmic}[1]
\Require tool proposal $p$, policy $\mathcal{P}$, runtime state $\Sigma$
\Ensure decision $d \in \{\textsc{allow}, \textsc{prompt}, \textsc{deny}\}$
\If{\Call{HardlineUnsafe}{$p$}}
    \State \Return \textsc{deny}
\EndIf
\If{$\Sigma.\mathrm{mode}=\texttt{read-only}$ and not $p.\mathrm{readOnly}$}
    \State \Return \textsc{deny}
\EndIf
\ForAll{$r \in \mathcal{P}.\mathrm{runDenyRules}$}
    \If{\Call{match}{$r,p,\Sigma$}}
        \State \Return \textsc{deny}
    \EndIf
\EndFor
\ForAll{$r \in \mathcal{P}.\mathrm{hostRules}$}
    \If{\Call{match}{$r,p,\Sigma$}}
        \State \Return $r.o$
    \EndIf
\EndFor
\ForAll{$r \in \mathcal{P}.\mathrm{modeRules}$}
    \If{\Call{match}{$r,p,\Sigma$}}
        \State \Return $r.o$
    \EndIf
\EndFor
\If{$p.\mathrm{requiresConfirmation}$}
    \State \Return \textsc{prompt}
\Else
    \State \Return \textsc{allow}
\EndIf
\end{algorithmic}
\end{algorithm}

The permission layer does not rely on natural-language instructions as a security boundary. Prompting may guide the model, but authority is enforced by runtime policy. This distinction is important because prompt injection, tool poisoning, and confused-deputy failures may cause a model to propose actions inconsistent with user intent. In Agentao, such proposals are still mediated by the ToolRunner pipeline and host-visible permission decisions.

\subsection{Tool System and Protocol-Mediated Capabilities}
\label{subsec:tool-system}

The tool system provides a uniform registration and invocation interface for heterogeneous capabilities. Capabilities may be built-in tools, host-injected tools, protocol-mediated tools, plugin-provided hooks, skills, or sub-agent wrappers. Each capability is represented by a name, schema, description, read-only flag, confirmation requirement, execution function, and optional session bindings such as working directory, filesystem abstraction, shell executor, or memory manager.

The runtime separates capability registration from capability authorization. Registration determines whether a capability is visible to the model and callable through the runtime. Authorization determines whether a particular invocation is permitted in the current context. This separation allows a host to reduce the model-visible tool surface using allowlists or disabled-tool settings while still relying on the permission engine as the enforcement boundary.

The core tool categories correspond to the tool system layer in Figure~\ref{fig:agentao-architecture}. File tools inspect, create, or modify local artifacts. Shell tools execute commands within the configured working directory and optional sandbox. Search tools provide repository and content search. Web tools access external information when enabled. Memory tools read or write scoped persistent state. Skill tools activate reusable task procedures. Agent tools delegate bounded tasks to sub-agents. User-query tools ask the user for clarification or confirmation. Planning, goal, and todo tools structure long-running tasks. MCP tools expose protocol-mediated external capabilities through the same governed invocation path.

Protocol-mediated tools are normalized into the same registry as local tools. This allows the runtime to apply the same permission decision function to a tool regardless of whether it originates from local code or an external protocol endpoint. Protocol composability therefore increases the action space without bypassing governance. The runtime can namespace externally derived tools, translate protocol schemas into model-visible tool schemas, and record their invocations in the same event trace as built-in tools.

\subsection{Memory, Skills, Plugins, and Sub-agents}
\label{subsec:memory-skills-plugins-subagents}

The supporting subsystems provide persistent state, reusable procedures, extension points, and multi-agent decomposition. These mechanisms improve long-horizon task performance, but they also expand the authority and context available to the agent. Agentao therefore treats them as governed runtime components rather than as hidden model-side features.

\paragraph{Memory.}
Memory is organized as scoped persistent state. A memory record can be associated with a user-level or project-level scope, allowing continuity without forcing all remembered information into a single global context. Memory retrieval is treated as a controlled state-access operation rather than an implicit model capability. For a user query $q$ and memory record $m$, the runtime computes a relevance score from interpretable signals:
\begin{equation}
\begin{aligned}
    S(m,q) ={}&
    w_{tag}\,\mathrm{Tag}(m,q)
    + w_{title}\,\mathrm{Title}(m,q)
    + w_{key}\,\mathrm{Keyword}(m,q) \\
    &+ w_{content}\,\mathrm{Content}(m,q)
    + w_{path}\,\mathrm{PathHint}(m,q)
    + w_{rec}\,\mathrm{Recency}(m)
    - w_{stale}\,\mathrm{Stale}(m).
\end{aligned}
\end{equation}
The explicit scoring terms make memory retrieval inspectable and debuggable. More importantly, scoped memory prevents long-term state from silently becoming unrestricted authority.

\paragraph{Skills.}
Skills provide reusable task procedures that can be discovered and activated across sessions. Architecturally, a skill contributes instructions or procedures to the agent context, but it does not bypass the governed tool pipeline. Concrete side effects still pass through permission checking and host-visible execution events.

\paragraph{Plugins and prompts.}
Plugins extend runtime behavior through lifecycle hooks. Pre-tool hooks can deny a proposed action, downgrade an allowed action to confirmation, or inject project-specific restrictions. Prompt hooks support context construction, compaction, and policy-aware prompt assembly. These hooks allow embedding applications to adapt runtime behavior without modifying the core execution loop.

\paragraph{Sub-agents.}
Sub-agents support task decomposition and specialization. A parent agent can delegate a bounded task to a child agent, either foregrounded or backgrounded depending on the host workflow. The runtime records sub-agent lineage so that the host can determine which parent session spawned which child task, whether the child completed, failed, or was cancelled, and how its result was incorporated. Sub-agent execution expands the runtime's coordination capacity, but it does not remove permission checks at the point where concrete tools are invoked.

\subsection{Observability, Replay, and Sessions}
\label{subsec:observability-replay-sessions}

Agentao exposes a structured host-facing event stream. The event stream projects internal runtime activity into typed, redacted events that can be consumed by an interactive interface, automation system, ACP transport, or embedded host. The most important event classes are permission-decision events, tool lifecycle events, sub-agent lifecycle events, transport updates, and replay records.

We model an execution trace as:
\begin{equation}
    \tau = \langle e_1, e_2, \ldots, e_m \rangle ,
\end{equation}
where each event has the form:
\begin{equation}
    e_i = \langle kind, session, turn, time, payload \rangle .
\end{equation}

For every attempted tool call $p$, the trace should contain a permission-decision event. For every allowed and executed tool call, the trace should also contain a start event and a terminal event. We express this requirement with the following trace invariant:
\begin{equation}
\label{eq:trace-invariant}
\forall p \in T_{\mathrm{exec}},\quad
\exists e_d,e_s,e_t \in \tau
\;\text{s.t.}\;
e_d \prec e_s \prec e_t .
\end{equation}
Here $e_d$ is a permission-decision event, $e_s$ is a tool-started event, and $e_t$ is a terminal event indicating either tool completion or tool failure. Denied and cancelled calls follow a different invariant: they must have a decision record and a terminal cancelled outcome, but they should not be represented to the host as ordinary successful tool executions.

Replay is implemented as an append-only event timeline. Each record is a structured event envelope, and schema snapshots are used to make replay files stable across releases. Replay serves three purposes. First, it supports debugging by allowing developers to reconstruct the sequence of model outputs, permission decisions, tool calls, and terminal outcomes. Second, it supports auditing by preserving the evidence needed to explain why a side effect occurred or did not occur. Third, it supports regression testing by allowing event traces to be compared against expected schemas and ordering constraints.

Session persistence complements replay. A session contains conversation state, task metadata, active permissions, memory references, and replay pointers. This allows long-running work to be paused, resumed, inspected, or embedded into host applications without reducing the traceability of the underlying execution.

\subsection{Execution Semantics and Concurrency}
\label{subsec:execution-semantics}

The execution layer supports both synchronous and asynchronous tools. Tool calls that survive planning, permission checking, and confirmation can be executed in batches. Independent tools may be executed in parallel, while calls to the same tool instance are serialized to protect shared callback or output state. 

Permission events are emitted before execution starts, and terminal lifecycle events are emitted when execution completes, fails, or is cancelled. Cancellation is treated as a first-class runtime state. A user or host may cancel a turn, a tool call, or a background sub-agent task. The runtime propagates cancellation tokens to tools that support them and emits terminal events so that the trace does not contain dangling starts without corresponding outcomes.

The model-visible conversation is updated only through formatted and sanitized tool result messages. This design preserves compatibility with strict tool-calling APIs that require every assistant tool call to receive a corresponding tool result. Even denied or cancelled calls are converted into explicit tool result messages that instruct the model not to retry the denied action through alternative means. This reduces the likelihood of a model repeatedly attempting to circumvent a policy decision.

\section{Case Analysis}
\label{sec:case-analysis}

This section presents three illustrative case studies that exercise
Agentao in concrete task workflows. The purpose of these cases is not to
provide a quantitative benchmark or to compare task success rates against
other agent frameworks. Instead, the goal is to show how the same runtime
harness can support governed tool use, scoped state, host-controlled
execution, protocol-composable capabilities, and structured
observability across different application settings.

Table~\ref{tab:case-mechanisms} summarizes the three cases. The first case
illustrates task-specific permission policy in a support-ticket workflow.
The second case illustrates local-first analytical execution over
workspace-scoped data and artifacts. The third case illustrates
unattended automation in which the runtime is driven by a scheduler
rather than by an interactive user.

\begin{table}
\centering
\small
\caption{Runtime mechanisms illustrated by the three case studies.}
\begin{tabular}{p{0.20\linewidth} p{0.34\linewidth} p{0.36\linewidth}}
\hline
\textbf{Case} & \textbf{Task-level decision point} & \textbf{Agentao mechanism exercised} \\
\hline
Support-ticket triage &
Whether a customer reply should be sent automatically or saved as a draft
for human review. &
Permission-engine customization, confidence-gated side effects,
custom tools, skill-shaped behavior, and host-visible tool events. \\

Data analysis workbench &
Whether local shell execution and artifact generation should be allowed
within a bounded analytical workspace. &
Workspace-scoped shell use, read-only data layout, skill composition,
artifact-return contract, and local-first execution. \\

Scheduled digest &
Whether an unattended run completed successfully and produced a
machine-consumable result. &
Headless runtime execution, iteration limits, strict output contract,
per-run state, event tapping, and fail-loud termination. \\
\hline
\end{tabular}

\label{tab:case-mechanisms}
\end{table}

\subsection{Support-Ticket Triage}
\label{subsec:case-ticket}

\begin{figure}[t]
\centering
\resizebox{\linewidth}{!}{%
\begin{tikzpicture}[
    font=\sffamily\scriptsize,
    >=Latex,
    stage/.style={
        draw,
        rounded corners=4pt,
        very thick,
        align=center,
        minimum height=0.95cm,
        inner sep=4pt
    },
    input/.style={
        stage,
        fill=blue!8,
        draw=blue!60!black,
        minimum width=2.8cm
    },
    tool/.style={
        stage,
        fill=teal!8,
        draw=teal!60!black,
        minimum width=3.2cm
    },
    draft/.style={
        stage,
        fill=violet!8,
        draw=violet!60!black,
        minimum width=3.2cm
    },
    allowbox/.style={
        stage,
        fill=green!10,
        draw=green!60!black,
        minimum width=3.3cm
    },
    reviewbox/.style={
        stage,
        fill=red!8,
        draw=red!65!black,
        minimum width=3.3cm
    },
    state/.style={
        stage,
        fill=gray!8,
        draw=gray!60!black,
        minimum width=3.5cm
    },
    eventbox/.style={
        stage,
        fill=yellow!18,
        draw=olive!55!black,
        minimum width=5.9cm,
        minimum height=1.0cm
    },
    policy/.style={
        draw=orange!85!black,
        fill=orange!15,
        very thick,
        diamond,
        aspect=2.2,
        align=center,
        inner sep=2.2pt,
        font=\sffamily\bfseries\scriptsize
    },
    panel/.style={
        draw,
        rounded corners=8pt,
        thick,
        inner sep=8pt
    },
    arrow/.style={
        -{Latex[length=2.2mm]},
        thick,
        draw=gray!75!black
    },
    auxarrow/.style={
        -{Latex[length=1.8mm]},
        thick,
        dashed,
        draw=gray!60!black
    },
    eventarrow/.style={
        -{Latex[length=1.8mm]},
        dotted,
        very thick,
        draw=magenta!70!black
    },
    policyallow/.style={
        -{Latex[length=2.2mm]},
        very thick,
        draw=green!55!black
    },
    policydeny/.style={
        -{Latex[length=2.2mm]},
        very thick,
        draw=red!70!black
    }
]

\node[input] (ticket) at (0,0) {Support\\ticket};

\node[tool] (profile) at (3.8,0)
    {\textsc{GetCustomerProfile}\\[-1pt]
     {\tiny read customer context}};

\node[tool] (kb) at (7.6,0)
    {\textsc{SearchKb}\\[-1pt]
     {\tiny retrieve support evidence}};

\node[draft] (draftreply) at (11.5,0)
    {\textsc{DraftReply}\\[-1pt]
     {\tiny no external side effect}};

\node[policy] (gate) at (15.4,0)
    {Policy\\gate};

\node[allowbox] (send) at (19.7,1.75)
    {\textsc{SendReply}\\[-1pt]
     {\tiny external side effect}};

\node[reviewbox] (human) at (19.7,-1.95)
    {Draft for\\human review};

\node[state] (skill) at (10.1,2.55)
    {Support-triage skill\\[-1pt]
     {\tiny tone, escalation, output contract}};

\node[state] (workspace) at (4.8,-2.25)
    {Per-ticket workspace\\[-1pt]
     {\tiny \texttt{runs/<ticket-id>/}}};

\node[eventbox] (events) at (15.8,-3.9)
    {HostEvent / replay trace\\[-1pt]
     {\tiny permission decision $\cdot$ tool lifecycle $\cdot$ terminal outcome}};

\draw[arrow] (ticket) -- (profile);
\draw[arrow] (profile) -- (kb);
\draw[arrow] (kb) -- (draftreply);
\draw[arrow] (draftreply) -- (gate);

\draw[policyallow]
    (gate.east) -- node[above, sloped, font=\sffamily\tiny]
    {allow: high confidence, low risk} (send.west);

\draw[policydeny]
    (gate.east) -- node[below, sloped, font=\sffamily\tiny]
    {prompt / deny} (human.west);

\draw[auxarrow] (skill.south west) -- (kb.north);
\draw[auxarrow] (skill.south east) -- (gate.north west);
\draw[auxarrow] (workspace.north) -- (profile.south);
\draw[auxarrow] (workspace.east) -- (draftreply.south west);

\draw[eventarrow] (draftreply.south) |- (events.west);
\draw[eventarrow] (gate.south) -- ++(0,-0.85) -| (events.north);
\draw[eventarrow] (send.south) |- (events.east);
\draw[eventarrow] (human.south) |- ([xshift=1.0cm]events.east);

\begin{scope}[on background layer]
\node[
    panel,
    fill=blue!3,
    draw=blue!30,
    fit=(ticket),
    label={[font=\sffamily\bfseries\scriptsize,text=blue!65!black]above:Input}
] {};

\node[
    panel,
    fill=teal!3,
    draw=teal!28,
    fit=(profile)(kb)(draftreply)(gate)(skill)(workspace),
    label={[font=\sffamily\bfseries\scriptsize,text=teal!60!black]above:Agentao governed runtime}
] {};

\node[
    panel,
    fill=gray!4,
    draw=gray!32,
    fit=(send)(human),
    label={[font=\sffamily\bfseries\scriptsize,text=gray!70!black]above:Host / external system}
] {};

\node[
    panel,
    fill=yellow!6,
    draw=olive!45!black,
    fit=(events),
    label={[font=\sffamily\bfseries\scriptsize,text=olive!55!black]below:Observability}
] {};
\end{scope}

\end{tikzpicture}%
}
\caption{Support-ticket triage workflow. Agentao embeds task-specific tools and a support-triage skill inside a governed runtime. The model may draft a reply, but the external side effect of sending it is mediated by a host-defined policy gate. Permission decisions, tool lifecycles, and terminal outcomes are emitted to the host-facing event stream and replay trace.}
\label{fig:ticket-case}
\end{figure}
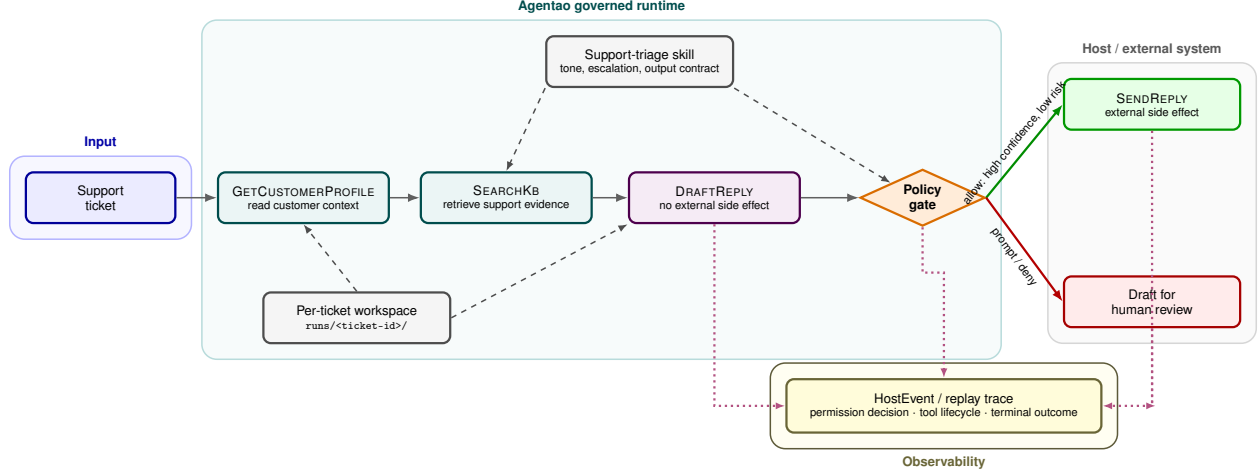

The first case considers a support-ticket triage workflow, as shown in
Fig.~\ref{fig:ticket-case}. A customer submits a ticket describing a
product issue, billing problem, or account request. The host embeds
Agentao as a task-specific runtime and provides custom tools for
retrieving customer context, searching a support knowledge base, drafting
a reply, and optionally sending the reply through an external support
system. The key point of the case is that the model may propose a
customer-facing action, but the side effect of sending a reply is
mediated by the runtime's permission policy rather than executed
directly.

A representative tool set is:
\[
T_{\mathrm{ticket}} =
\{\textsc{GetCustomerProfile}, \textsc{SearchKb},
\textsc{DraftReply}, \textsc{SendReply}\}.
\]
The first two tools are read-oriented capabilities. They allow the agent
to inspect customer metadata and retrieve relevant documentation. The
\textsc{DraftReply} tool produces a candidate response without external
side effects. The \textsc{SendReply} tool, however, commits an external
action: it sends a message to the customer or updates the ticket system.
The workflow therefore separates low-risk information gathering from
high-impact communication.

Agentao's governance layer allows the host to express this distinction
as policy rather than as a natural-language instruction alone. For
example, the host may allow \textsc{GetCustomerProfile},
\textsc{SearchKb}, and \textsc{DraftReply} in a workspace-write or
task-specific mode, while routing \textsc{SendReply} through an
additional confidence and confirmation policy. Abstractly, the host
policy can be written as:
\[
D(\textsc{SendReply}, a, \Sigma, \mathcal{P}) =
\begin{cases}
\textsc{allow}, & a.\mathrm{confidence} \geq 0.9,\\
\textsc{deny}, & \text{otherwise}.
\end{cases}
\]
Here \(a\) denotes the structured tool arguments proposed by the model,
\(\Sigma\) is runtime state, \(P\) is the active policy, and
\(\theta > \theta'\) are host-selected thresholds. The important point is
not the specific thresholding rule, but the separation of model judgment
from host authorization. The model may propose that a reply should be
sent, but the runtime determines whether this side effect is allowed,
requires human confirmation, or must be denied.

A typical execution proceeds as follows. The agent first reads the ticket
and calls \textsc{GetCustomerProfile} to obtain account context. It then
calls \textsc{SearchKb} to retrieve relevant support articles. A
task-specific skill shapes the response format, tone, escalation policy,
and required evidence. The agent produces a candidate response through
\textsc{DraftReply}. If the response is low risk and the host-defined
policy permits automation, the agent may call \textsc{SendReply}. If the
policy requires review, Agentao emits a permission request and the host
or human operator decides whether to approve the action.

This case illustrates that governance can be task-specific rather than
only mode-specific. The decision to send a reply is not delegated solely
to the LLM; it is mediated by a host-defined permission policy over
structured tool arguments, active runtime state, and the side-effect
profile of the tool. The host-facing event stream records the permission
decision and the lifecycle of each tool call, making it possible to
inspect which customer context was requested, which knowledge-base search
was performed, whether a reply was drafted or sent, and whether human
approval was required.

\subsection{Data Analysis Workbench}
\label{subsec:case-data-workbench}

The second case considers a local data analysis workbench. A user asks a natural-language question about a local dataset, such as a sales table, application log, or collection of parquet files. The host embeds Agentao in a workspace that exposes the dataset through a local data view and provides a writable working area for generated queries, intermediate files, notes, and charts. The agent is allowed to inspect schemas, write analysis scripts, execute local analytical commands, and generate artifacts, but it should not be granted unrestricted authority over the host environment.

A representative workspace contains a data view and generated artifacts:
\[
W =
\{\texttt{data/},\ \texttt{cache-*},\ \texttt{chart-*},\ \texttt{.agentao/skills/}\}.
\]
In the runnable example, the shared dataset is exposed through a \texttt{data/} link inside a per-session workspace, while reusable procedures are provided through local skills. A production deployment may additionally mount \texttt{data/} as read-only and route generated artifacts to a dedicated artifact directory.

The active tool set may include file-reading tools, file-writing tools scoped to the workspace, shell execution, search, and skills for DuckDB-based analysis and matplotlib chart generation. The agent can therefore transform a natural-language question into a concrete analytical workflow:
\[
q_{\mathrm{user}}
\rightarrow
\text{schema inspection}
\rightarrow
\text{query generation}
\rightarrow
\text{local execution}
\rightarrow
\text{artifact generation}
\rightarrow
\text{final explanation}.
\]

The key runtime property is that analytical execution is tool-mediated and workspace-scoped. The agent may propose shell commands such as running DuckDB or executing a Python plotting script, but these commands pass through the permission-mediated tool pipeline before execution. The host can configure the runtime to allow ordinary analytical commands inside the workspace, prompt for broad shell commands, and deny attempts to modify source data or access paths outside the intended project boundary.

A typical run begins with schema inspection. The agent lists available tables or parquet files, reads a small sample or schema summary, and constructs a SQL query. It then writes the query or script into the workspace and executes it through a shell-mediated local command. If the result should be visualized, the agent writes a plotting script and generates a chart artifact. The runnable example uses an application-level artifact marker of the following form:
\begin{quote}
\small
\texttt{[CHART] chart-<slug>.png}
\end{quote}
This marker gives the host a simple contract for identifying generated chart artifacts in the model-visible final response or streamed text output.

This case illustrates how Agentao supports useful local computation without granting unrestricted ambient authority. The runtime does not need to send the dataset to a remote service, and the model does not directly execute arbitrary commands against the host environment. Instead, the host exposes a bounded analytical capability surface, while Agentao mediates command execution, file writes, artifact generation, and event exposure. When the host subscribes to the event stream or enables replay, the same runtime interface can be used to inspect which files were read, which commands were executed, which artifacts were created, and whether any operation required confirmation or was denied.

\subsection{Unattended Scheduled Digest}
\label{subsec:case-scheduled-digest}

The third case considers an unattended scheduled digest. In this setting, Agentao is not driven by an interactive user. Instead, a scheduler such as \texttt{cron}, a workflow engine, or a containerized batch job launches the runtime at a fixed time, provides a task prompt, and expects a structured result. Example tasks include summarizing daily project changes, producing a nightly issue digest, compiling a status report, or checking a workspace for newly generated artifacts.

This case differs from the previous two because there may be no human present to answer clarification questions or approve permission prompts during execution. The host therefore configures the runtime with explicit iteration limits, a constrained tool surface, a fixed working directory, and a strict output contract. A representative prompt contract is:
\begin{quote}
\small
Write \texttt{digest.md} in the working directory and end the final assistant message with exactly one line of the form:
\begin{verbatim}
RESULT: {"path": "digest.md", "items": N}
\end{verbatim}
\end{quote}
Here \texttt{N} denotes the number of digest items produced. The important property is that the final model response contains a machine-parseable terminal marker that the scheduler can validate.

A typical execution proceeds as follows. The scheduler creates a per-run workspace, for example:
\begin{quote}
\small
\texttt{runs/YYYY-MM-DD/}
\end{quote}
Agentao receives a task prompt and runs in a non-interactive mode with a bounded number of iterations. The agent reads the configured sources, writes \texttt{digest.md}, and terminates with the required \texttt{RESULT:} line. The runner then parses this line and emits a scheduler-facing JSON object, for example:
\begin{quote}
\small
\begin{verbatim}
{"status": "ok", "date": "YYYY-MM-DD",
 "tokens_est": 1234, "path": "digest.md", "items": N}
\end{verbatim}
\end{quote}
This separates the model-facing output contract from the outer automation contract consumed by the scheduler.

If the agent cannot complete the task, encounters a denied operation, omits the required terminal marker, or exceeds the iteration budget, the runtime fails loudly. The scheduler can then retry, alert an operator, or preserve the event stream and replay artifacts for later inspection when such recording is enabled.

This case illustrates Agentao's use as a headless runtime harness. In an interactive assistant, a permission prompt can be resolved by the user. In an unattended batch workflow, the host must instead decide in advance which capabilities are available and how failures should be surfaced. Agentao supports this pattern by making active permissions, iteration limits, tool outcomes, session state, and host-visible events explicit runtime artifacts. The result is an automation workflow whose authority boundary is defined by the host rather than improvised by the model at run time.

\section{Limitations}
\label{sec:limitations}

Although Agentao illustrates a governed runtime design for local-first LLM agents, the system has several limitations. We highlight three that are particularly important for interpreting the scope of the paper.

\paragraph{No formal guarantee of agent safety.}
The runtime constrains agent behavior through permission modes, policy rules, confirmation mechanisms, and structured execution boundaries, but these mechanisms do not constitute a formal proof of safety. The system can reduce the authority of model-mediated actions and make risky operations more visible, but it cannot guarantee that all unsafe behavior will be prevented. Its effectiveness depends on the correctness of policy configuration, tool metadata, host enforcement, and the underlying operating-system mechanisms. Therefore, Agentao should be understood as a practical governance layer rather than a formally verified security monitor.

\paragraph{Dependence on capability descriptions and host-side enforcement.}
The governance model assumes that capabilities can be meaningfully described, classified, and mediated before execution. In practice, tool descriptions, plugin metadata, protocol schemas, and host-provided abstractions may be incomplete, stale, ambiguous, or adversarially manipulated. A runtime can reject unknown tools, enforce permission modes, and record decisions, but it cannot fully infer the semantic consequences of every external capability. Similarly, protocol-mediated tools and third-party extensions may introduce side effects outside the runtime's direct control. This limitation suggests that governed runtimes should be combined with capability provenance, sandboxing, supply-chain review, and deployment-specific policy validation.

\paragraph{Absence of quantitative evaluation.}
This paper focuses on the architecture and design rationale of a governed local-first agent runtime, but does not provide a quantitative benchmark evaluation. As a result, it does not measure permission-enforcement accuracy, runtime overhead, protocol compatibility at scale, or task-utility trade-offs against existing agent frameworks. Such evaluation is necessary to establish the empirical effectiveness and cost of the proposed runtime design, and remains future work.

\section{Conclusion}
This paper presented Agentao as a governed local-first runtime for LLM agents. The central claim is that tool-using agents should be understood not only as model-driven planners, but also as execution systems whose actions require explicit authority boundaries, inspectable state transitions, and recoverable traces. From this perspective, runtime governance is complementary to model alignment, prompting strategies, and workflow orchestration.

We characterized the design space through a motivation and threat model centered on over-privileged actions, prompt injection, tool poisoning, state leakage, and weak auditability. We then described a layered architecture that separates model-generated action proposals from host-authorized execution through a host contract, a runtime core, a permission-mediated tool pipeline, scoped state, protocol-composable capabilities, and a structured observability interface.

Agentao does not eliminate the risks of autonomous tool use, nor does it provide formal safety guarantees. Its contribution is more modest: it demonstrates how a practical agent runtime can make permissions, state, protocols, and execution traces explicit system abstractions. We argue that such runtime-level abstractions are necessary for building LLM agents that are not merely capable, but also governable, inspectable, and suitable for integration into local and host-controlled environments.

\section*{Ethical Considerations}

Agentao exposes powerful tool-using capabilities that may affect local files, persistent memory, and external services. While the runtime is designed around explicit permissions, confirmation, scoped state, and structured audit traces, these mechanisms do not eliminate the risks of misuse, unsafe automation, or accidental side effects. Deployers should grant only the minimum necessary authority, review third-party tools and protocol servers before use, protect replay logs and memory stores as sensitive data, and maintain appropriate human oversight for high-impact actions.

\bibliographystyle{unsrt}  
\bibliography{references}  

\end{document}